 \documentclass[journal,10pt,twocolumn]{IEEEtran}
\usepackage{setspace}
\usepackage{booktabs}
\usepackage[dvips]{epsfig}    
\usepackage{enumerate}
\usepackage{graphicx}         
\usepackage{remreset}
\usepackage{subfigure}
\usepackage{float}

\makeatletter \@removefromreset{figure}{section} \makeatother
\usepackage{enumerate, color}
\usepackage{amssymb}
\usepackage{amsmath}
\usepackage{framed}
\usepackage{mathrsfs}
\usepackage{amsmath,bm}
\usepackage{amsthm}
\usepackage{multirow}
\usepackage{array}
\usepackage{verbatim}
\usepackage{color}
\usepackage{subfigure}
\usepackage{multicol}
\usepackage[section]{placeins}
\usepackage{footnote}

\begin{document}
\title{GXNOR-Net: Training deep neural networks with ternary weights and activations without full-precision memory under a unified discretization framework}

%

 \author{Lei Deng${^*}$, Peng Jiao${^*}$, Jing Pei${^\dag}$,  Zhenzhi Wu and  Guoqi Li${^\dag}$\\
\par{(Please cite us with:L. Deng, et al. GXNOR-Net: Training deep neural networks with ternary weights and activations without full-precision memory under a unified discretization framework. Neural Networks 100, 49-58(2018).)}
 \thanks{L. Deng,  P. Jiao, J. Pei, Z. Wu and G. Li  are with the
 Department  of Precision Instrument, Center for Brain Inspired Computing Research, Tsinghua University, Beijing, China, 100084. L. Deng is also with the Department of Electrical and Computer Engineering, University of California, Santa Barbara, CA 93106, USA.  Emails: leideng@ucsb.edu (L. Deng), jiaop15@mails.tsinghua.edu.cn (P. Jiao), peij@mail.tsinghua.edu.cn (J. Pei),  wuzhenzhi@mail.tsinghua.edu.cn (Z. Wu) and liguoqi@mail.tsinghua.edu.cn (G. Li).
 ${^*}$  L. Deng and P. Jiao contribute equally to this work.
$^\dag$  The corresponding authors: Guoqi Li and Jing Pei.} }  


\maketitle                       

 \begin{abstract}
 Although deep neural networks (DNNs) are being a revolutionary power to open up the  AI era, the notoriously huge hardware overhead has  challenged their applications. Recently, several  binary and ternary networks,   in which  the costly multiply-accumulate operations   can be replaced   by accumulations or even  binary logic operations,   make the on-chip training of DNNs quite promising. Therefore there is a pressing need to build an  architecture  that could subsume   these networks  under  a unified framework that  achieves both  higher performance and less overhead. To this end,   two fundamental issues are yet to be addressed. The first one is how to implement the back propagation when neuronal activations are discrete.  The second one is  how to remove  the full-precision hidden weights in the training phase  to break the bottlenecks of memory/computation consumption.  To address  the first issue, we present  a  multi-step neuronal  activation discretization method and    a  derivative approximation   technique  that enable the implementing  the back propagation algorithm on discrete DNNs. While for the second issue, we propose a discrete state transition (DST)  methodology   to constrain the weights in a discrete space without saving the hidden weights.  Through this way, we  build a unified framework that  subsumes  the binary or ternary networks as  its special cases,   and  under which a heuristic algorithm is provided   at the website https://github.com/AcrossV/Gated-XNOR. More particularly, we find that when both  the weights and activations  become  ternary  values, the DNNs can be reduced to sparse binary networks, termed as gated XNOR  networks  (GXNOR-Nets) since only  the event of non-zero weight and non-zero activation enables the control gate  to start the XNOR logic operations in the original binary networks. This promises the event-driven hardware design for efficient mobile intelligence. We achieve advanced performance compared with state-of-the-art algorithms. Furthermore, the computational  sparsity and  the number of states  in the  discrete space can be flexibly modified to make it suitable for various hardware platforms.

\end{abstract}


\section{Introduction}

Deep neural networks (DNNs) are rapidly developing with the use of big data sets, powerful models/tricks and GPUs, and have been widely  applied in various fields \cite{Vision 1}-\cite{Multi_modal 1}, such as vision, speech, natural language, Go game, multimodel tasks, etc. However, the huge hardware overhead is also notorious, such as  enormous memory/computation resources and high power consumption, which has greatly challenged their applications. As we know, most of the DNNs  computing overheads  result  from the costly multiplication of real-valued synaptic weight and real-valued neuronal activation, as well as the accumulation operations.   Therefore, a few compression methods and binary/ternary networks emerge in recent years, which aim to put DNNs on efficient devices. The former ones \cite{Compression 1}-\cite{Compression 6} reduce the network parameters and connections, but most of them do not change the full-precision multiplications and accumulations. The latter ones \cite{BWN_Bengio 1}-\cite{XNOR 2016}  replace the original computations  by only accumulations or even  binary logic operations.

In particular, the binary weight networks (BWNs) \cite{BWN_Bengio 1}-\cite{BWN/TWN_CAS 2016} and ternary weight networks (TWNs) \cite{BWN/TWN_CAS 2016} \cite{TWN_Han 2016} constrain the synaptic weights to the binary space $\{-1,1\}$ or the ternary space $\{-1,0,1\}$, respectively. In this way, the multiplication operations can be removed. The binary neural networks (BNNs) \cite{BNN_Bengio 2016} \cite{XNOR 2016} constrain both the synaptic weights and the neuronal activations to the binary space $\{-1,1\}$, which can directly replace the multiply-accumulate operations by binary logic operations, i.e. XNOR. So this kind of networks is also called  the XNOR networks. Even with these most advanced models, there  are  issues that remain  unsolved.  Firstly, the reported networks are based on specially designed discretization and training methods,  and there is a pressing need to build an  architecture  that could subsume   these networks under  a  unified framework that  achieves both  higher performance and less overhead.  To this end,  how to implement the back propagation for online training algorithms  when the activations are constrained in a discrete space is yet to be  addressed. On the other side, in all these networks we have to save the full-precision hidden weights in the training phase, which causes frequent data exchange between the external memory for parameter storage and internal buffer for forward and backward computation.

In this paper, we propose a  discretization framework: (1) A multi-step discretization function that constrains the neuronal activations in a discrete space, and a method to implement the back propagation  by introducing an approximated derivative for the  non-differentiable activation function; (2) A discrete state transition (DST)  methodology  with a  probabilistic projection  operator which constrains the synaptic weights in a discrete space without the storage of full-precision hidden weights in the whole training phase.  Under  such a discretization  framework, a heuristic algorithm is provided   at the website https://github.com/AcrossV/Gated-XNOR, where the state number of weights and activations are reconfigurable to make it suitable for various hardware platforms. In the  extreme case, both the weights and activations can be  constrained in the ternary space $\{-1,0,1\}$ to form ternary neural networks (TNNs). For a multiplication operation, when one of the weight and activation is zero or both of them are zeros, the corresponding computation unit is resting, until the non-zero weight and non-zero activation enable and wake up the required computation unit. In other words, the computation trigger determined by the weight and activation acts as a control signal/gate  or an event to start the computation. Therefore,  in contrast to the existing XNOR networks, the TNNs proposed in this paper can be treated as gated XNOR networks (GXNOR-Nets). We test this network model over MNIST, CIFAR10 and SVHN datasets, and achieve comparable performance with state-of-the-art  algorithms. The efficient hardware architecture is designed and compared with conventional ones. Furthermore, the sparsity of the neuronal activations can be flexibly modified to improve the recognition performance and hardware efficiency. In short, the GXNOR-Net promises the ultra efficient hardware for future mobile intelligence based on the reduced memory and computation, especially for the event-driven running paradigm.

 We define several abbreviated terms that will be used in the following sections: (1)CWS: continuous weight space; (2)DWS: discrete weight space; (3)TWS: ternary weight space; (4)BWS: binary weight space; (5)CAS: continuous activation space; (6)DAS: discrete activation space; (7)TAS: ternary weight space; (8)BAS: binary activation space; (9)DST: discrete state transition.

\section{Unified discretization framework  with   multi-level states of  synaptic weights and neuronal activations in DNNs}
Suppose that there are  $K$ training samples   given by  $\{(x^{(1)}, y^{(1)}), ... (x^{(\kappa)}, y^{(\kappa)}), ...,  (x^{(K)}, y^{(K)})\}$ where $y^{(\kappa)}$ is the label of the $\kappa$th sample  $x^{(\kappa)}$.  In this work,
we are going to propose  a general  deep architecture  to efficiently  train DNNs  in which both the synaptic weights and neuronal activations are  restricted  in a discrete space  $Z_N$  defined as
\begin{equation}
    \begin{array}{lll}
    \large
Z_N=\left\{z^n_N| z^n_N=(\frac{n}{2^{N-1}}-1),~n=0, 1, ... ,2^N\right\}
    \end{array}
    \label{ZN}
    \end{equation}
    where    $N$ is  a given non-negative integer, i.e., $N=0, 1, 2, ... $ and   $\Delta z_N=\frac{1}{2^{N-1}}$   is  the distance between adjacent states. \bigskip

 {\it {\textbf{Remark 1.} } Note that different values of $N$   in    $Z_N$  denote different discrete spaces.  Specifically, when $N=0$, $Z_N=\{-1, 1\}$ belongs to  the  binary space and $\Delta z_0=2$.  When $N=1$, $Z_N=\{-1, 0,  1\}$ belongs to  the  ternary  space and   $\Delta z_1=1$.
 Also  as seen in (\ref{ZN}),  the  states in  $Z_N$  are  constrained
in the interval $[-1, 1]$, and without loss of generality,  the range  can be easily extended to $[-H, H]$   by multiplying a scaling factor $H$.  } \bigskip

In the following subsections,  we first investigate the  problem formulation for  GXNOR-Nets, i.e,  $Z_N$ is constrained in the
ternary space $\{-1,0,1\}$. Later we will   investigate  how to implement back  propagation in DNNs  with ternary synaptic weights and neuronal activations. Finally  a unified discretization framework by extending the weights and activations to multi-level states will be presented.

\subsection{Problem formulation for  GXNOR-Net}
\begin{figure}[!htbp]
\begin{center}
\includegraphics[height=5.5cm]{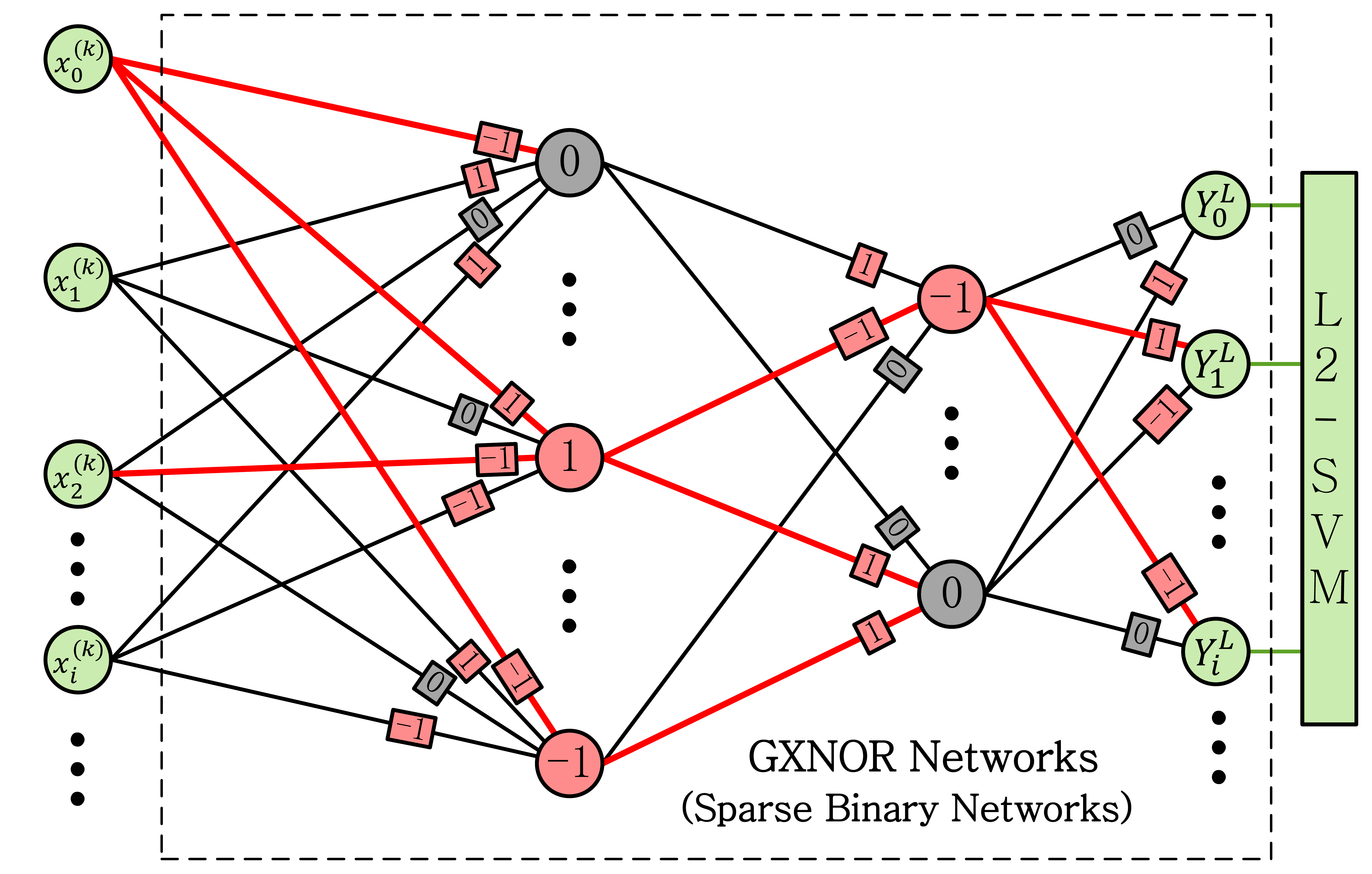}
\caption{\textbf{GXNOR-Net.} In a GXNOR-Net, when both the pre-neuronal activation and synaptic weight are non-zero, the forward computation is required, marked as red. This indicates the GXNOR-Net is a sparse binary network, and most of the computation units can be switched off which reduces power consumption. The enable signal determined by the corresponding weight and activation acts as a control gate for the computation.}
\label{TNN}
\end{center}
\end{figure}

By constraining  both the  synaptic weights  and neuronal activations  to binary  states $\{ -1, 1 \}$  for the computation
 in both forward and backward passes, the complicated  float multiplications and accumulations  change to
be very simple logic operations  such as XNOR. However, different from XNOR networks,   GXNOR-Net can be regarded as a sparse binary network   due to the existence of the zero state,  in which the number of zero state   reflects the networks'  sparsity.   Only when both the pre-neuronal activation and synaptic weight are non-zero, the forward computation is required, marked as red as seen in  Fig. \ref{TNN}.  This indicates that  most of the computation resources can be switched off to reduce  power consumption. The enable signal determined by the corresponding weight and activation acts as a control gate for the computation. Therefore, such a network  is called  the gated XNOR network (GXNOR-Net). Actually, the sparsity is also leveraged by other neural networks, such as in \cite{Knoblauch 2010} \cite{Knoblauch 2016}.

Suppose that there are $L+1$  layers in a   GXNOR-Net where both the synaptic weights and neuronal activations are  restricted  in a discrete space  $Z_1=\{-1,0,1\}$
except the  zeroth  input layer and the activations of the $L$th layer.  As shown   in Fig. \ref{TNN}, the last  layer, i.e., the $L$th layer is   followed by a $L_2$-SVM output layer with the standard hinge loss, which has been shown to perform better than  softmax on several benchmarks \cite{SVM1 2013}\cite{SVM2 2015}.

Denote  $Y^{l}_i$  as   the activation of neuron $i$ in layer $l$   given by
  \begin{equation}
 Y^{l}_i=\varphi\left(\sum_j W^{l}_{ij}Y^{l-1}_j\right)
   \end{equation}
 for  $1 \leq l \leq L-1$, where $\varphi(.)$ denotes an  activation function and $W^{l}_{ij}$ represents the synaptic weight between neuron $j$ in layer $l-1$ and neuron $i$ in layer $l$.  For the $\kappa$th training sample,    $Y_i^{0}$  represents  the $i$th element of the  input vector of $x^{(\kappa)}$, i.e.,  $Y_i^{0}=x_i^{(\kappa)}\in R$.
For the  $L$th layer  of GXNOR-Net    connected with   the L2-SVM output layer,  the neuronal activation $Y_i^L\in R$.

  The optimization model of  GXNOR-Net is formulated as follows
\begin{equation}
\begin{array}{lll}
&argmin_{W,Y} ~~~~  E(W,Y)    \\
&s.t.  ~~ W^{l}_{ij}\in \{-1, 0, 1\}, Y^{l}_i\in \{-1, 0, 1\} ~~l=1, 2 ...., L-1\\
&~~~~~~ Y^{l}_i=\varphi(\sum_j W^{l}_{ij}Y^{l-1}_j),~~l=1, 2 ...., L-1\\
&~~~~~~Y^{l}_i=\sum_j W^{l}_{ij}Y^{l-1}_j,~~l=L\\
&~~~~~~Y_i^{l}=x_i^{(\kappa)},~~ l=0, ~\kappa=1, 2, ... , K
\label{optimizationmodelbe}
\end{array}
\end{equation}
  Here $E(W,Y)$   represents  the cost function  depending on  all synaptic weights (denoted as $W$) and neuronal activations (denoted as $Y$) in all layers  of  the   GXNOR-Net.

For the convenience of presentation, we  denote the discrete space  when  describing  the synaptic weight and  the neuronal activation as  the DWS   and  DAS, respectively. Then,  the  special ternary space  for  synaptic weight and neuronal activation  become  the respective TWS and  TAS. Both  TWS and TAS are the  ternary   space $Z_1=\{-1, 0, 1\}$ defined in (1).

The objective is to  minimize  the cost function $E(.)$  in   GXNOR-Nets  by constraining  all the synaptic weights and  neuronal activations  in TWS and TAS  for  both forward and backward passes.   In the  forward  pass,  we will first  investigate  how to discretize the  neuronal activations   by introducing  a quantized  activation  function. In the   backward   pass,  we will  discuss how to implement the back propagation with ternary neuronal activations through approximating  the derivative  of the   non-differentiable   activation  function. After that, the  DST   methodology for weight update  aiming to solve (\ref{optimizationmodelbe}) will be presented.

\subsection{Ternary neuronal activation discretization  in the  forward  pass}
 We   introduce   a   quantization function $\varphi_{r}(x)$
 to  discretize  the neuronal activations  $Y^{l}$ ($1\leq l \leq L-1$) by setting
\begin{equation}
 \begin{array}{lll}
Y_i^{l}=\varphi_r \left(\sum_j W_{ij}^lY_j^{l-1}\right)
 \label{optimizationmodelbe2}
 \end{array}
    \end{equation}
where
   \begin{equation}
\varphi_{r}(x) = \begin{cases}
1,  &\mbox{if   $x> r$} \\
0,   &\mbox{if   $|x|\leq r$}\\
-1,   &\mbox{if   $x< -r$}\\
\end{cases}
\label{activityfunction}
   \end{equation}
   \begin{figure}[!htbp]
\begin{center}
\includegraphics[height=8.8cm]{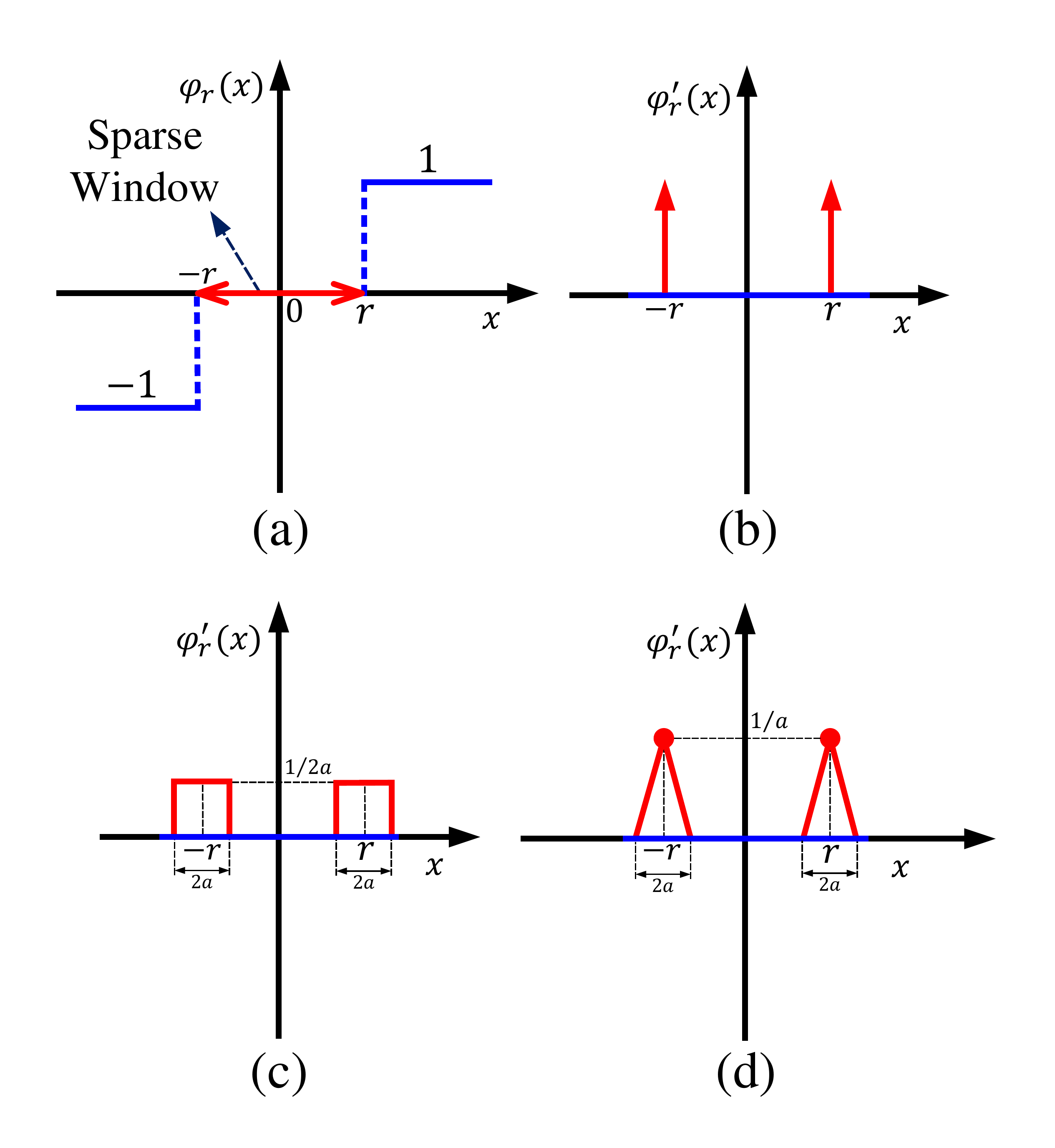}
\caption{\textbf{Ternary discretization of neuronal activations  and derivative approximation methods.} The quantization function $\varphi_{r}(x)$ (a) together with its ideal derivative in (b) can be approximated by (c) or (d).}
\label{neuron_discrete1}
\end{center}
\end{figure}
In Fig. \ref{neuron_discrete1}, it  is seen that  $\varphi_{r}(x)$ quantizes the neuronal activation to  the TAS $Z_1$  and $r>0$ is a window  parameter which controls the excitability of the  neuron and the sparsity of the computation.

\subsection{Back propagation with ternary neuronal activations through approximating  the derivative  of the quantized  activation  function}
After the  ternary neuronal activation discretization in the forward pass,     model (\ref{optimizationmodelbe})  has now  been simplified to the following optimization model
\begin{equation}
\begin{array}{lll}
&argmin_{W} ~~~~  E(W)    \\
&s.t. ~~~~   W^{l}_{ij}\in \{-1, 0, 1\}, ~ ~l=1,2,...,L\\
&~~~~~~~~ Y^{l}_i=\varphi_{r}(\sum_j W^{l}_{ij}Y^{l-1}_j),  ~ ~l=1, 2 ...., L-1\\
&~~~~~~~~ Y^{l}_i=\sum_j W^{l}_{ij}Y^{l-1}_j,~ ~l=L\\
&~~~~~~~~Y_i^{0}=x_i^{(\kappa)},  ~~ \kappa=1, 2, ... , K
\label{optimizationmodelbe1}
\end{array}
\end{equation}

As mentioned in the Introduction section, in order to implement the back propagation  in the  backward  pass  where  the neuronal activations are discrete, we need to obtain the derivative of the quantization function $\varphi_{r}(x)$ in  (\ref{activityfunction}).  However, it is well known that  $\varphi_{r}(x)$ is not continuous and non-differentiable, as shown in Fig. \ref{neuron_discrete1}(a) and (b). This makes it difficult to implement the back propagation in  GXNOR-Net in this case.  To  address this issue, we approximate the derivative of  $\varphi_{r}(x)$ with respect to $x$  as follows
 \begin{equation}
\frac{\partial \varphi_{r}(x)}{\partial x} =
\begin{cases}
\frac{1}{2a}, &\mbox{if $ r-a\leq |x|  \leq  r+a$}\\
0, &others
\end{cases}
\label{approcimation1}
\end{equation}
where $a$ is a small positive parameter  representing the
steep degree of the derivative in the neighbourhood of $x$. In real applications, there are many other ways to approximate the  derivative. For example, $\frac{\partial \varphi_{r}(x)}{\partial x}$ can also be approximated as
 \begin{equation}
\frac{\partial \varphi_{r}(x)}{\partial x} =
\begin{cases}
- \frac{1}{a^2} \left(|x|-(r+a)\right), &\mbox{if $ r\leq |x|   \leq  r+a$}\\
 \frac{1}{a^2} \left(|x|-(r-a)\right), &\mbox{ if $r-a\leq |x| < r$}  \\
0, &others
\end{cases}
\label{approcimation}
\end{equation}
 for a small given parameter $a$.
The above two approximated methods are shown in Fig. \ref{neuron_discrete1}(c) and (d), respectively. It is seen that  when $a \rightarrow 0$, $\frac{\partial \varphi_{r}(x)}{\partial x}$ approaches  the impulse function in Fig. \ref{neuron_discrete1}(b).

  Note that the  real-valued  increment  of  the synaptic weight $W^{l}_{ij}$ at the  $k$th iteration at layer $l$, denoted as   $\Delta W^{l}_{ij}(k)$, can be   obtained  based on   the gradient information£º
    \begin{equation}
  \Delta W^{l}_{ij}(k)=-\eta   \cdot   \frac{\partial E\left(W(k), Y(k)\right)}{\partial W^{l}_{ij}(k)}
  \label{gradient}
    \end{equation}
    where $\eta$¡¡ represents  the learning rate  parameter,  $W(k)$ and   $Y(k)$ denote  the  respective synaptic weights and neuronal activations  of  all layers at the  current  iteration,  and
      \begin{equation}
 \frac{\partial E\left(W(k), Y(k)\right)}{\partial W^{l}_{ij}(k)}=  Y_j^{l-1}   \cdot    \frac{\partial \varphi_{r}(x^l_i) }{ \partial  x^l_i  }       \cdot  e^{l}_i
    \end{equation}
    where   $x^l_i$ is a  weighted sum of  the  neuron $i$'s  inputs  from layer $l-1$:
      \begin{equation}
  x^{{l}}_i= \sum_{j} W^l_{ij}  Y_{j}^{l-1}
    \end{equation}
    and  $e^{l}_i$ is the error signal  of  neuron $i$   propagated from layer $l+1$:
    \begin{equation}
e^{l}_i=  \sum_{\iota} W^{l+1}_{\iota i}   \cdot  e^{l+1}_{\iota} \cdot    \frac{\partial  \varphi_{r}(x_\iota^{l+1}) }{ \partial  x_\iota^{l+1}  }
    \end{equation}
    and both $ \frac{\partial \varphi_{r}(x^l_i) }{ \partial  x^l_i  } $ and $ \frac{\partial \varphi_{r}(x_\iota^{l+1}) }{ \partial  x_\iota^{l+1}  } $ are  approximated through (\ref{approcimation}) or (\ref{approcimation1}). As mentioned,  the $L$th layer is followed by the L2-SVM output layer, and the   hinge  foss  function \cite{SVM1 2013}\cite{SVM2 2015} is applied for the training. Then, the  error back   propagates from the output layer to anterior layers and the gradient information for each layer can be obtained accordingly.

    \bigskip
 \subsection{Weight update by discrete state transition in the ternary weight space}
Now we investigate  how to solve (\ref{optimizationmodelbe1}) by constraining  $W$  in the  TWS through  an iterative training process. Let  $W^{l}_{ij}(k) \in Z_1$   be   the weight state at the $k$-th iteration step, and  $\Delta W^{l}_{ij}(k)$ be the  weight increment on $W^{l}_{ij}(k)$ that can be  derived on the gradient information (\ref{gradient}).  To guarantee the next weight will not jump out of $[-1,1]$,  define   $\varrho (\cdot)$  to establish a boundary restriction  on  $\Delta W^{l}_{ij}(k)$:
\small
  \begin{equation}
  \varrho (\Delta W^l_{ij}(k)) =
    \begin{cases}
    min\left(1-W^l_{ij}(k),   \Delta W^l_{ij}(k) \right) &\mbox{if $\Delta W^l_{ij}(k)\geq 0$}\\
    max\left(-1-W^l_{ij}(k),   \Delta W^l_{ij}(k) \right) &\mbox{else}
    \end{cases}
    \end{equation} \normalsize
 and  decompose the above $\varrho (\Delta W^{l}_{ij}(k))$ as:
  \begin{equation}
    \begin{array}{lll}
  \varrho (\Delta W^l_{ij}(k))  =   \kappa_{ij}\Delta  z_1+ \nu_{ij}=   \kappa_{ij} + \nu_{ij}
    \end{array}
    \end{equation}
such that
  \begin{equation}
    \begin{array}{lll}
  \kappa_{ij}= fix\left( \varrho (\Delta W^l_{ij}(k))/\Delta z_1\right)= fix\left( \varrho (\Delta W^l_{ij}(k))\right)
    \end{array}
    \end{equation}
    and
 \begin{equation}
    \begin{array}{lll}
 \nu_{ij}=rem\left(\varrho (\Delta W^l_{ij}(k)), \Delta z_1\right)= rem\left(\varrho (\Delta W^l_{ij}(k)), 1\right)
    \end{array}
    \end{equation}
 where $fix(.)$  is a round operation towards zero, and  $rem(x,y)$ generates the remainder of the division between two numbers and keeps the same sign with $x$.

Then, we  obtain a  projected weight increment  $\Delta w_{ij}(k)$  and update the weight by
   \begin{equation}
    \begin{array}{lll}
    W^l_{ij}(k+1)&=   W^l_{ij}(k)+   \Delta w_{ij}(k)\\
    &=W^l_{ij}(k)+ \mathcal{P}_{grad}\left(\varrho (\Delta W^l_{ij}(k))\right)
     \end{array}
     \label{projectionupdate}
     \end{equation}
     Now we discuss how to project  $\Delta w_{ij}(k)$ in CWS to make the next state $W^l_{ij}(k)+ \mathcal{P}_{grad}\left(\varrho (\Delta W^l_{ij}(k))\right)$ in TWS, i.e. $W^l_{ij}(k+1) \in Z_N$.  We  denote $\Delta w_{ij}(k)=\mathcal{P}_{grad}(.)$ as    a  probabilistic projection function    given by
 \begin{equation}
    \begin{array}{lll}
   {P} \left(  \Delta  w_{ij}(k)   =\kappa_{ij}\Delta  z_1+ sign(\varrho (\Delta W^l_{ij}(k))) \Delta  z_1 \right) =\mathbf{\tau}(\nu_{ij}) \\
{P} \left(   \Delta  w_{ij}(k) = \kappa_{ij}\Delta  z_1 \right) = 1-\mathbf{\tau}(\nu_{ij}) \\
    \end{array}
    \label{projectionprobality}
    \end{equation}
    where the sign function  $sign(x)$  is given by
\begin{equation}
sign(x) =
\begin{cases}
1, &\mbox{if $x \geq 0$}\\
-1, &\mbox{else}
\end{cases}
\end{equation}
   and  $\mathbf{\tau}(.)$   ($0\leq \mathbf{\tau}(.) \leq 1$)  is  a state transition probability  function defined by
   \begin{equation}
 \mathbf{\tau}(\nu)=
     tanh\left( m \cdot \frac{|\nu|}{\Delta z_N} \right)   
  \label{nonlinear_pro}
  \end{equation}
   where $m$ is a nonlinear factor of positive constant to adjust  the transition probability in probabilistic projection.

 The above formula (\ref{projectionprobality}) implies that $\Delta w_{ij}(k)$
 is among   $\kappa_{ij} +1$,   $\kappa_{ij} -1$ and
    $\kappa_{ij}$.  For example, when   $sign(\varrho (\Delta W^l_{ij}(k)))=1$,
then   $\Delta  w_{ij}(k)=\kappa_{ij}+ 1$ happens with   probability  $\mathbf{\tau}(\nu_{ij})$ and $   \Delta  w_{ij}(k) = \kappa_{ij}$ happens with   probability  $1-\mathbf{\tau}(\nu_{ij})$.
Basically the  $ \mathcal{P}_{grad}(.)$  describes  the transition  operation among discrete states in $Z_1$ defined in  (1), i.e., $ Z_1=\{z^n_1| z^n_1=n-1, n=0, 1, 2\}$ where $z_1^0=-1$,  $z_1^1=0$ and $z_1^2=1$.

\begin{figure}[!htbp]
\begin{center}
\includegraphics[height=5.7cm]{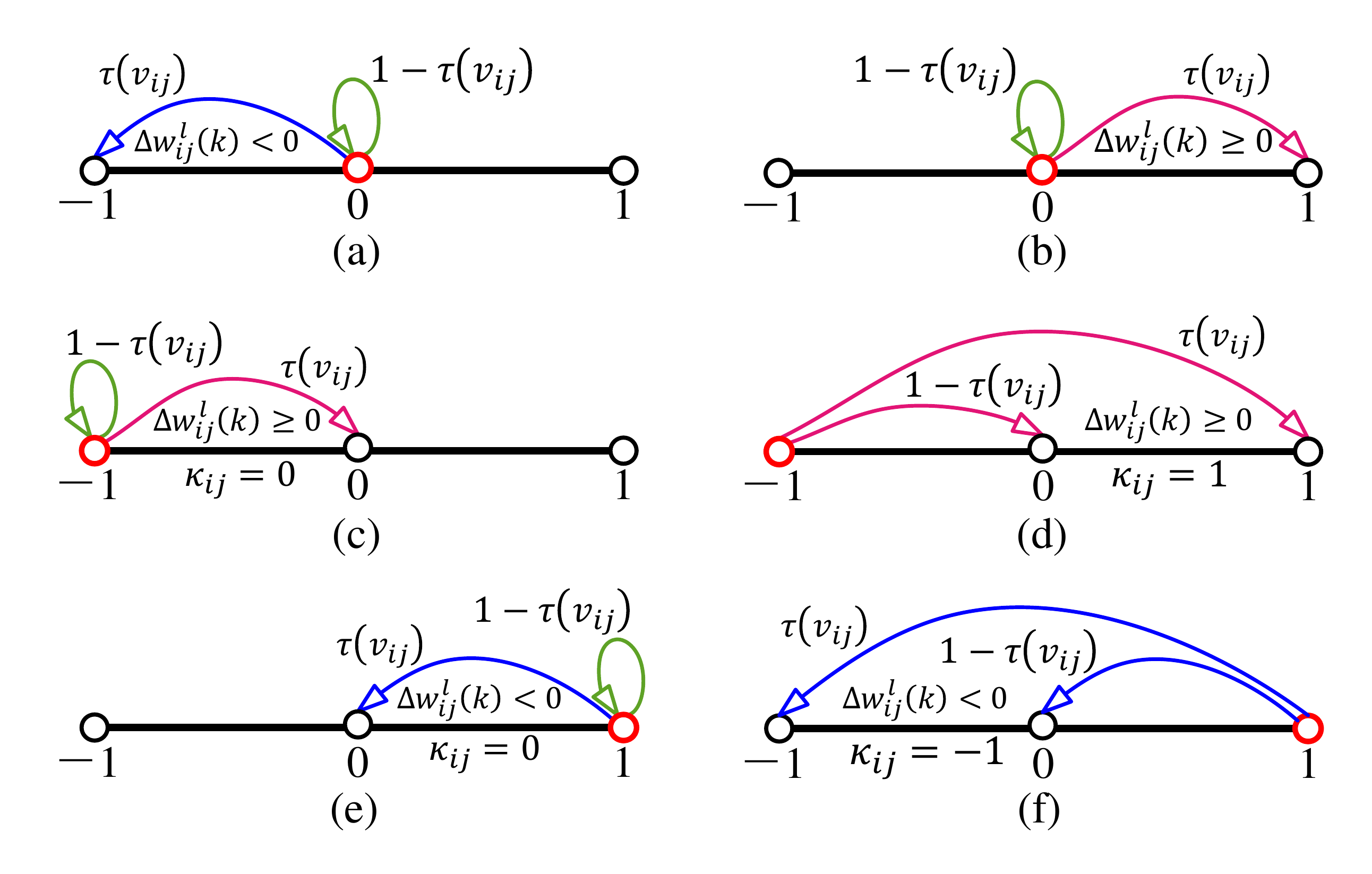}
\caption{\textbf{Illustration of DST in TWS.}  In DST, the weight can directly transit from current discrete state (marked as red circle) to the next discrete state when updating the weight, without the storage of the full-precision hidden weight. With different current weight states, as well as the direction and magnitude of weight increment $\Delta W_{ij}^{l}(k)$, there are totally six transition cases when the discrete space is the TWS.}
\label{weight_discrete2}
\end{center}
\end{figure}

Fig. \ref{weight_discrete2}   illustrates the transition process in TWS.  For example, at   the current weight state  $W^l_{ij}(k)=z_1^1=0$,
if  $\Delta W^l_{ij}(k)< 0$, then $W^l_{ij}(k+1)$ has the probability of $\mathbf{\tau}(\nu_{ij})$ to transfer to $z_1^0=-1$ and has the probability of $1-\mathbf{\tau}(\nu_{ij})$  to stay at $z_1^1=0$; while if   $\Delta W^l_{ij}(k)\geq 0$, then $W^l_{ij}(k+1)$ has the probability of $\mathbf{\tau}(\nu_{ij})$ to transfer to $z_1^2=1$ and has the probability of $1-\mathbf{\tau}(\nu_{ij})$  to stay at $z_1^1=0$.
 At the boundary state  $W^l_{ij}(k)=z_1^0=-1$,     if $\Delta W^l_{ij}(k) < 0$, then $\varrho (\Delta W^l_{ij}(k))=0$ and
${P} \left(  \Delta w   =0 \right)= 1$,   which means that  $W^l_{ij}(k+1)$    has the probability of $1$  to stay at $z_1^0=-1$; if $\Delta W^l_{ij}(k)\geq 0$ and $\kappa_{ij}=0$, ${P} \left(  \Delta w   = 1 \right)=  \mathbf{\tau}(\nu_{ij})$, then
$W^l_{ij}(k+1)$ has the probability of $\mathbf{\tau}(\nu_{ij})$ to transfer to $z_1^1=0$, and  has the probability of $1-\mathbf{\tau}(\nu_{ij})$  to stay at $z_1^0=-1$;
if $\Delta W^l_{ij}(k)\geq 0$ and $\kappa_{ij}=1$, ${P} \left(  \Delta w   = 2 \right)=  \mathbf{\tau}(\nu_{ij})$, then
$W^l_{ij}(k+1)$ has the probability of $\mathbf{\tau}(\nu_{ij})$ to transfer to $z_1^2=1$, and  has the probability of $1-\mathbf{\tau}(\nu_{ij})$  to transfer to $z_1^1=0$. Similar analysis holds  for another boundary state  $W^l_{ij}(k)=z_1^2=1$.

\begin{figure}[!htbp]
\begin{center}
\includegraphics[height=4.2cm]{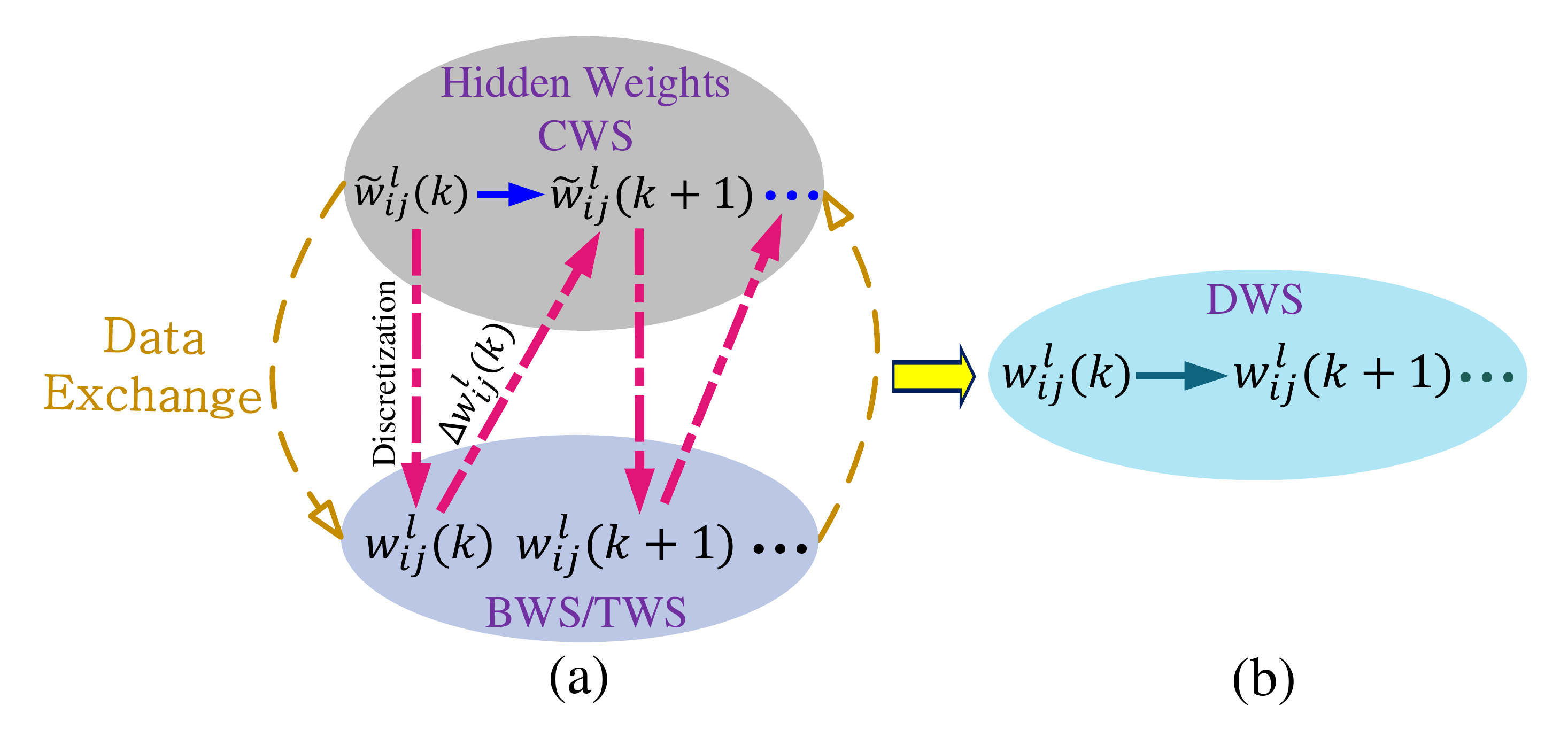}
\caption{\textbf{Illustration of the discretization of synaptic weights.}  (a)  shows that existing schemes frequently switch between two spaces, i.e., CWS and BSW/TWS at each iteration. (b) shows that the weights by using our DST are always constrained in DWS during the whole training phase.}
\label{weight_discrete1}
\end{center}
\end{figure}

Based on the above results,  now we  can  solve the optimization model (\ref{optimizationmodelbe}) based on  the DST methodology. The main idea is to update the  synaptic weight based on (\ref{projectionupdate})  in the ternary space $Z_1$ by exploiting the projected gradient information.  The main difference  between DST and the ideas in recent works such as  BWNs \cite{BWN_Bengio 1}-\cite{BWN/TWN_CAS 2016},  TWNs \cite{BWN/TWN_CAS 2016} \cite{TWN_Han 2016}, BNNs or  XNOR networks  \cite{BNN_Bengio 2016} \cite{XNOR 2016}  is illustrated in Fig. \ref{weight_discrete1}. In those works, frequent switch and data exchange between the CWS and the BWS or TWS  are required during the training phase. The full-precision weights have to be saved at each iteration, and the gradient computation is based on the binary/ternary  version of the stored full-precision weights, termed as ``binarization"  or ``ternary discretization" step.  In stark contrast, the weights in  DST are always constrained in a DWS.    A probabilistic gradient projection operator   is introduced in (\ref{projectionprobality}) to directly transform a continuous weight increment to a discrete state transition.

{\it \textbf{Remark 2.} In the inference phase, since  both the synaptic weights and neuronal activations are in the ternary space,  only logic operations are required. In the training phase, the remove of full-precision hidden weights drastically  reduces  the memory cost. The logic forward pass and additive backward pass (just a bit of multiplications at each neuron node) will also simplify the training computation to some extent. In addition, the number of zero state, i.e. sparsity, can be controlled by adjusting   $r$ in $\varphi_r(.)$, which further makes our framework efficient in real applications through the event-driven paradigm.}

 \subsection{Unified discretization framework: multi-level states of the synaptic weights  and neuronal activations}
Actually, the binary and ternary  networks are not  the whole story since  $N$   is   not limited to be $0$ or $1$ in $Z_N$  defined in (\ref{ZN})  and  it can be  any non-negative integer.  There are many hardware platforms that support multi-level discrete space for more powerful processing ability \cite{TrueNorth 2014}-\cite{Memristor 2015}.

\begin{figure}[!htbp]
\begin{center}
\includegraphics[height=5.5cm]{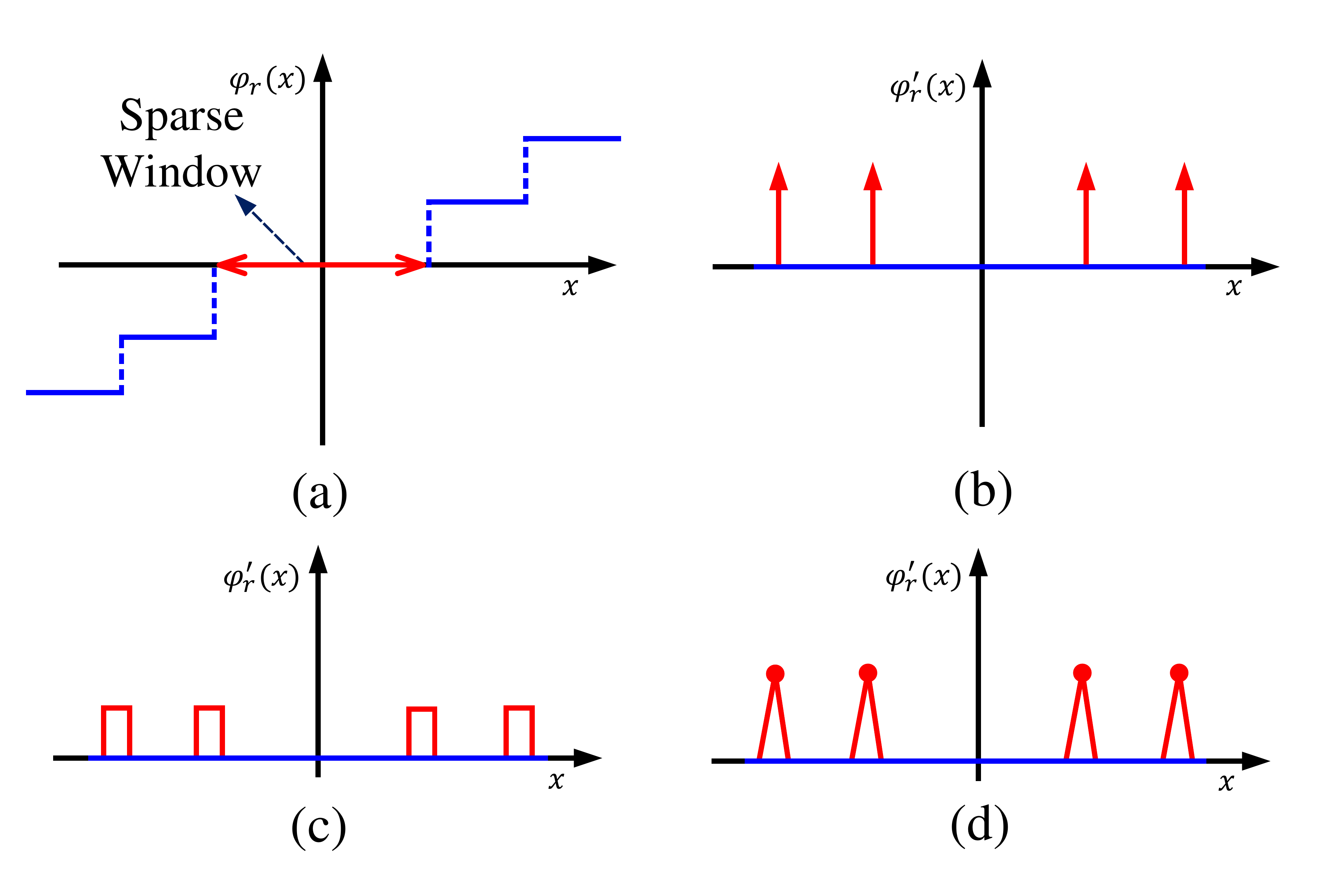}
\caption{\textbf{Discretization of neuronal activations with multi-level values and derivative approximation methods.} The multiple level of the quantization function $\varphi_{r}(x)$ (a) together with its ideal derivative in (b) can be approximated by (c) or (d).}
\label{neuron_discrete2}
\end{center}
\end{figure}

The neuronal activations can be extended to multi-level cases. To this end, we introduce the following  multi-step neuronal activation discretization function
\begin{equation}
 \begin{array}{lll}
Y_i^{l}=\varphi_r \left(\sum_j W_{ij}^{l}Y_j^{l-1}\right)
 \end{array}
    \end{equation}
   where
   \small
 \begin{equation}
\varphi_{r}(x) = \begin{cases}
\frac{\omega}{2^{N-1}},  &\mbox{if   $\frac{\omega-1}{2^{N-1}} (H-r)\leq x-r \leq  \frac{\omega}{2^{N-1}}(H-r)$}  \\
0,   &\mbox{if   $|x|< r$}\\
-\frac{\omega}{2^{N-1}},   &\mbox{if   $-\frac{\omega}{2^{N-1}} (H-r)\leq x+r \leq   -\frac{\omega-1}{2^{N-1}}(H-r)$} \\
\end{cases}
   \end{equation} \normalsize
for $ 1\leq \omega \leq 2^{N-1}$. The interval $[-H,H]$ is similarly defined with $Z_N$ in (\ref{ZN}). To implement the back  propagation  algorithm, the derivative of $\varphi_{r}(x)$ can be approximated at each discontinuous point as illustrated in Fig. \ref{neuron_discrete2}. Thus, both the forward pass and backward pass of DNNs can be implemented.

\begin{figure}[!htbp]
\begin{center}
\includegraphics[height=2.9cm]{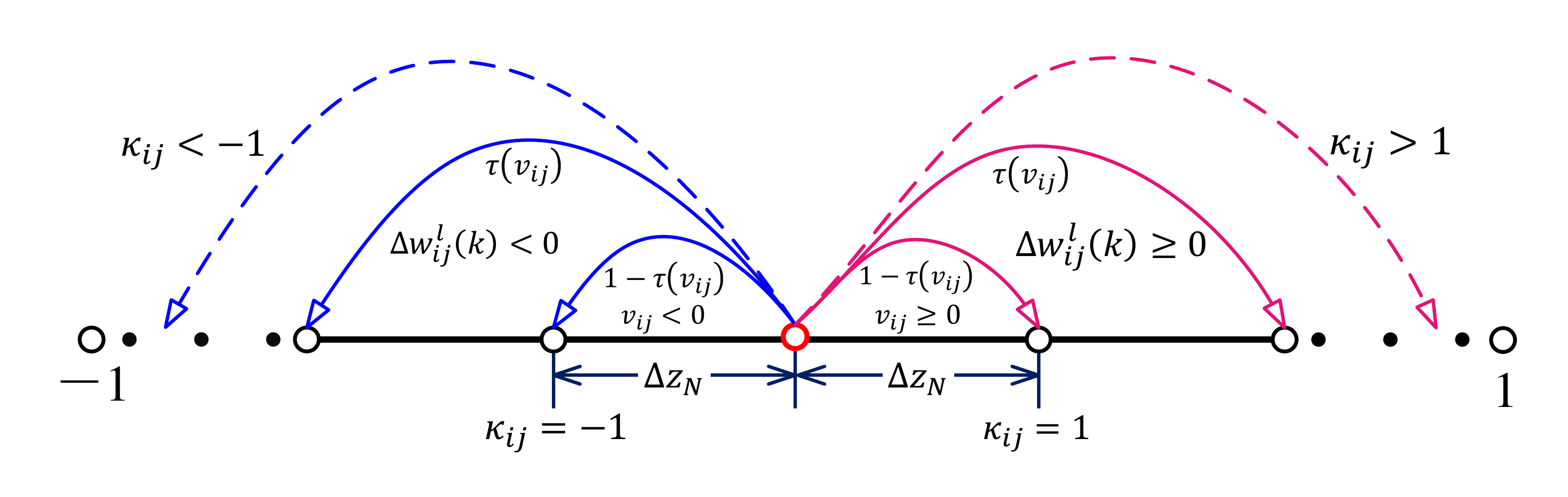}
\caption{\textbf{Discretization of synaptic weights in DWS with multi-level states.}  }
\label{weight_discrete3}
\end{center}
\end{figure}

At the same time,the proposed  DST for weight update can also be implemented in a discrete  space with multi-level states. In this case, the  decomposition of   $\Delta W_{ij}^l(k)$  is revisited as
  \begin{equation}
    \begin{array}{lll}
  \varrho (\Delta W_{ij}^l(k))  =   \kappa_{ij}\Delta z_N+ \nu_{ij}
    \end{array}
    \end{equation}
such that
  \begin{equation}
    \begin{array}{lll}
  \kappa_{ij}= fix\left( \varrho (\Delta W_{ij}^l(k))/\Delta z_N\right)
    \end{array}
    \end{equation}
    and
 \begin{equation}
    \begin{array}{lll}
 \nu_{ij}=rem\left(\varrho (\Delta W_{ij}^l(k)), \Delta z_N\right)
    \end{array}
    \end{equation}
and the   probabilistic projection function  in (\ref{projectionprobality}) can also be revisited  as follows
 \begin{equation}
    \begin{array}{lll}
   {P} \left(  \Delta w_{ij}(k)   =\kappa_{ij} \Delta z_N  + sign(\varrho (\Delta W^l_{ij}(k))) \Delta z_N \right) =\mathbf{\tau}(\nu_{ij}) \\
{P} \left(   \Delta w_{ij}(k)  = \kappa_{ij} \Delta z_N \right) = 1-\mathbf{\tau}(\nu_{ij}) \\
    \end{array}
    \end{equation}
 Fig. \ref{weight_discrete3}  illustrates  the state transition  of synaptic weights in DWS. In contrast to the transition example of TWS in Fig. \ref{weight_discrete2}, the  $\kappa_{ij}$ can be larger than $1$ so that further transition is allowable.

\section{Results}

We test the proposed GXNOR-Nets over the MNIST, CIFAR10 and SVHN datasets\footnote{\textbf{The  codes are available at https://github.com/AcrossV/Gated-XNOR}}. The results are shown in Table \ref{performance}. The network structure for MNIST is ``32C5-MP2-64C5-MP2-512FC-SVM'', and that for CIFAR10 and SVHN is ``2$\times$(128C3)-MP2-2$\times$(256C3)-MP2-2$\times$(512C3)-MP2-1024FC-SVM''. Here $MP$, $C$ and $FC$ stand for max pooling, convolution and full connection, respectively. Specifically, $2\times (128C3)$ denotes 2 convolution layers with $3\times3$ kernel and 128 feature maps, MP2 means max pooling with window size $2\times 2$ and stride 2, and 1024FC represents a full-connected layer with 1024 neurons.  Here SVM is a classifier with squared hinge loss (L2-Support Vector Machine)  right after the output layer.   All the inputs are normalized into the range of [-1,+1]. As for CIFAR10 and SVHN, we adopt the similar augmentation in \cite{SVM2 2015}, i.e. 4 pixels are padded on each side of training images, and a $32\times 32$ crop is further randomly sampled from the padded image and its horizontal flip version. In the inference phase, we only test using the single view of the original $32\times 32$ images. The batch size over MNIST, CIFAR10, SVHN are $100$, $1000$ and $1000$, respectively. Inspired by \cite{BNN_Bengio 2016}, the learning rate decays at each training epoch by $LR=\alpha \cdot LR$, where $\alpha$ is the decay factor determined by $\sqrt[Epochs]{LR\_fin/LR\_start}$. Here $LR\_start$ and $LR\_fin$ are the initial and final learning rate, respectively, and $Epochs$ is the number of total training epochs. The transition probability factor in equation (\ref{nonlinear_pro}) satisfies $m=3$, the derivative approximation uses rectangular window in Fig. \ref{neuron_discrete1}(c) where $a=0.5$. The base algorithm for gradient descent is Adam, and the presented performance is the accuracy on testing set.

\subsection{Performance comparison}

\small
\begin{table}
\centering
\renewcommand\thetable{\arabic{table}}
\caption{\textbf{Comparisons with state-of-the-art algorithms and networks.}}
\begin{tabular}{ccccccccccc}
\hline
\multirow{2}*{Methods} &\multicolumn{3}{c}{Datasets}   \\
&\cline{1-3}
 &MNIST &CIFAR10 &SVHN  \\
\hline
BNNs \cite{BNN_Bengio 2016} 	&98.60\% 	&89.85\%  &97.20\%   \\
\hline
TWNs \cite{BWN/TWN_CAS 2016}  	&99.35\% 	&92.56\%  &N.A  \\
\hline
BWNs \cite{BWN_Bengio 2}  &98.82\%  	&91.73\%   &97.70\%  \\
\hline
BWNs \cite{BWN/TWN_CAS 2016}  &99.05\%  	&90.18\%   &N.A   \\
\hline
Full-precision NNs \cite{BWN/TWN_CAS 2016}  	&99.41\% 	&92.88\%  &N.A    \\
\hline
\textbf{GXNOR-Nets}  &\textbf{99.32\%} 	&\textbf{92.50\%}  &\textbf{97.37\%}  \\
\hline
\label{performance}
\end{tabular}
\end{table}
\normalsize

\begin{figure}[!htbp]
\begin{center}
\includegraphics[height=4.6cm]{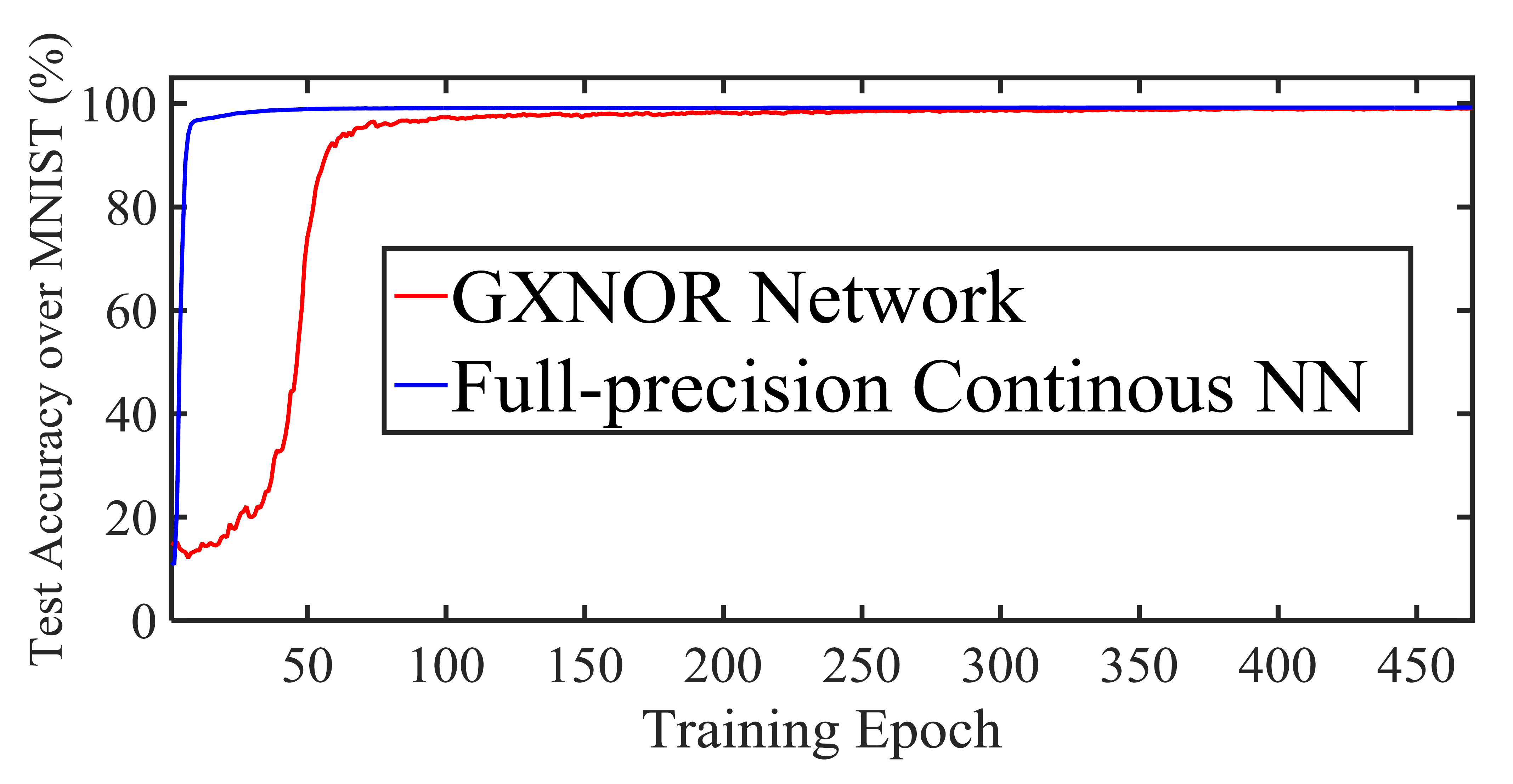}
\caption{\textbf{Training curve.} GXNOR-Net can achieve comparable final accuracy, but converges  slower than full-precision continuous NN.}
\label{training_curve}
\end{center}
\end{figure}

The networks for comparison in Table \ref{performance} are listed as follows:  GXNOR-Nets in this paper (ternary synaptic weights and ternary neuronal activations), BNNs or XNOR networks (binary synaptic weights and binary neuronal activations), TWNs (ternary synaptic weights and full-precision neuronal activations), BWNs (binary synaptic weights and full-precision neuronal activations), full-precision NNs (full-precision synaptic weights and full-precision neuronal activations). Over MNIST, BWNs \cite{BWN_Bengio 2} use full-connected networks with 3 hidden layers of 1024 neurons and a L2-SVM output layer, BNNs \cite{BNN_Bengio 2016} use full-connected networks with 3 hidden layers of 4096 neurons and a L2-SVM output layer, while our paper adopts the same structure as BWNs \cite{BWN/TWN_CAS 2016}. Over CIFAR10 and SVHN, we remove the last full-connected layer in BWNs \cite{BWN_Bengio 2} and BNNs \cite{BNN_Bengio 2016}. Compared with BWNs \cite{BWN/TWN_CAS 2016}, we just replace the softmax output layer by a L2-SVM layer. It is seen that  the proposed  GXNOR-Nets achieve comparable performance with  the state-of-the-art  algorithms and networks. In fact, the accuracy of 99.32\% (MNIST), 92.50\% (CIFAR10) and 97.37\% (SVHN) has outperformed most of the existing binary or ternary methods. In  GXNOR-Nets, the weights are always constrained in the TWS $\{-1,0,1\}$ without saving the full-precision hidden weights like the reported networks in Table \ref{performance}, and the neuronal activations are further constrained in the TAS $\{-1,0,1\}$. The  results   indicate that it is really possible to perform well even if we just use this kind of extremely hardware-friendly network architecture. Furthermore, Fig. \ref{training_curve} presents the graph where the error curve evolves as a function of the training epoch. We can see that the GXNOR-Net can achieve comparable final accuracy, but converges    slower than full-precision continuous NN.

\subsection{Influence of $m$, $a$ and $r$}

\begin{figure}[!htbp]
\begin{center}
\includegraphics[height=4.4cm]{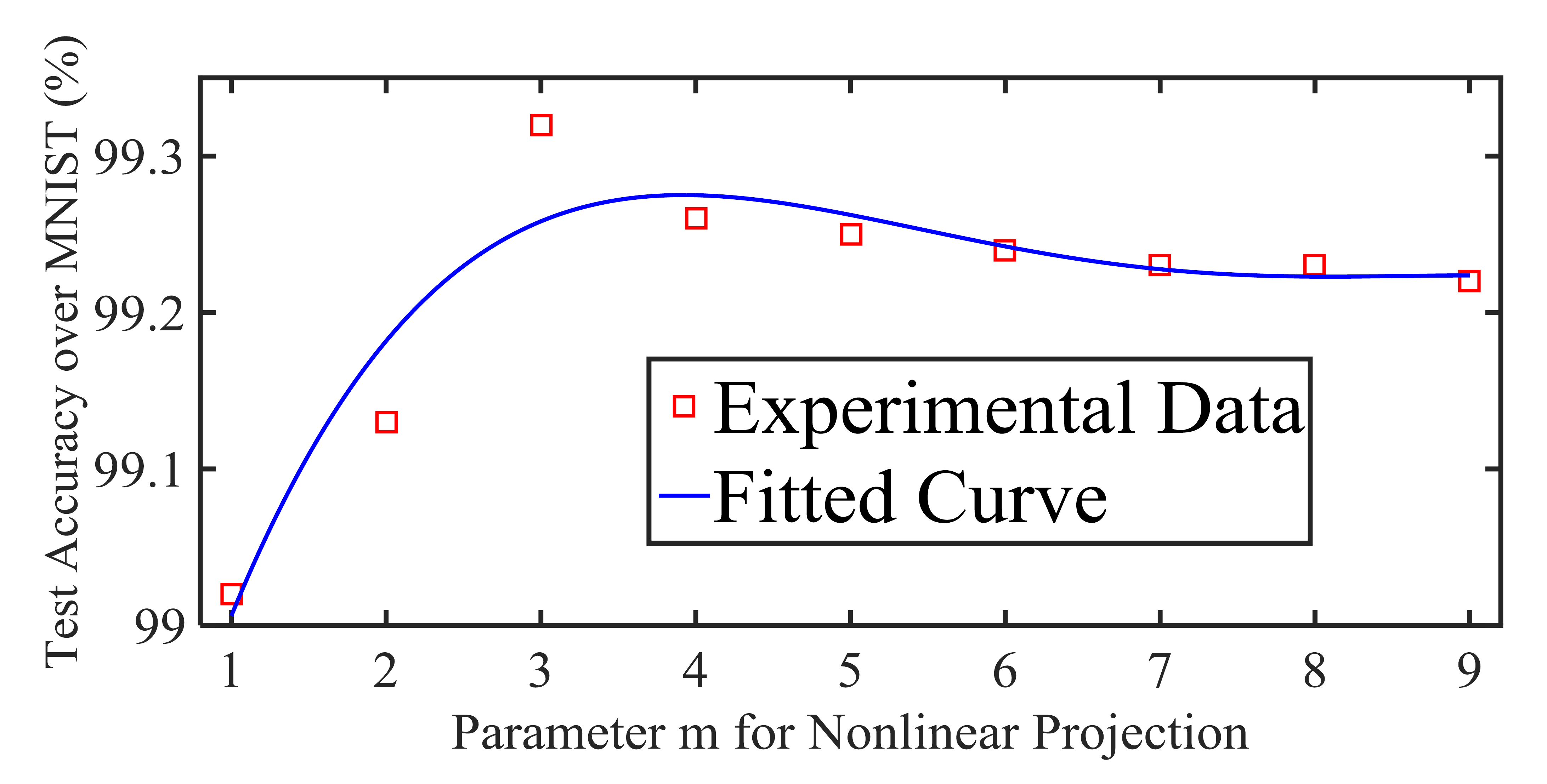}
\caption{\textbf{Influence of the nonlinear factor for probabilistic projection.} A properly larger $m$ value obviously improves performance, while too large value further helps little.}
\label{tanh_m}
\end{center}
\end{figure}

We  analyze the influence of several parameters in this section. Firstly, we study the nonlinear factor $m$ in equation (\ref{nonlinear_pro}) for probabilistic projection. The results are shown in Fig. \ref{tanh_m}, in which larger $m$ indicates stronger nonlinearity. It is seen that properly increasing $m$ would obviously improve the network performance, while too large $m$ further helps little. $m=3$ obtains the best accuracy, that is the reason why we use this value for other experiments.

\begin{figure}[!htbp]
\begin{center}
\includegraphics[height=4.4cm]{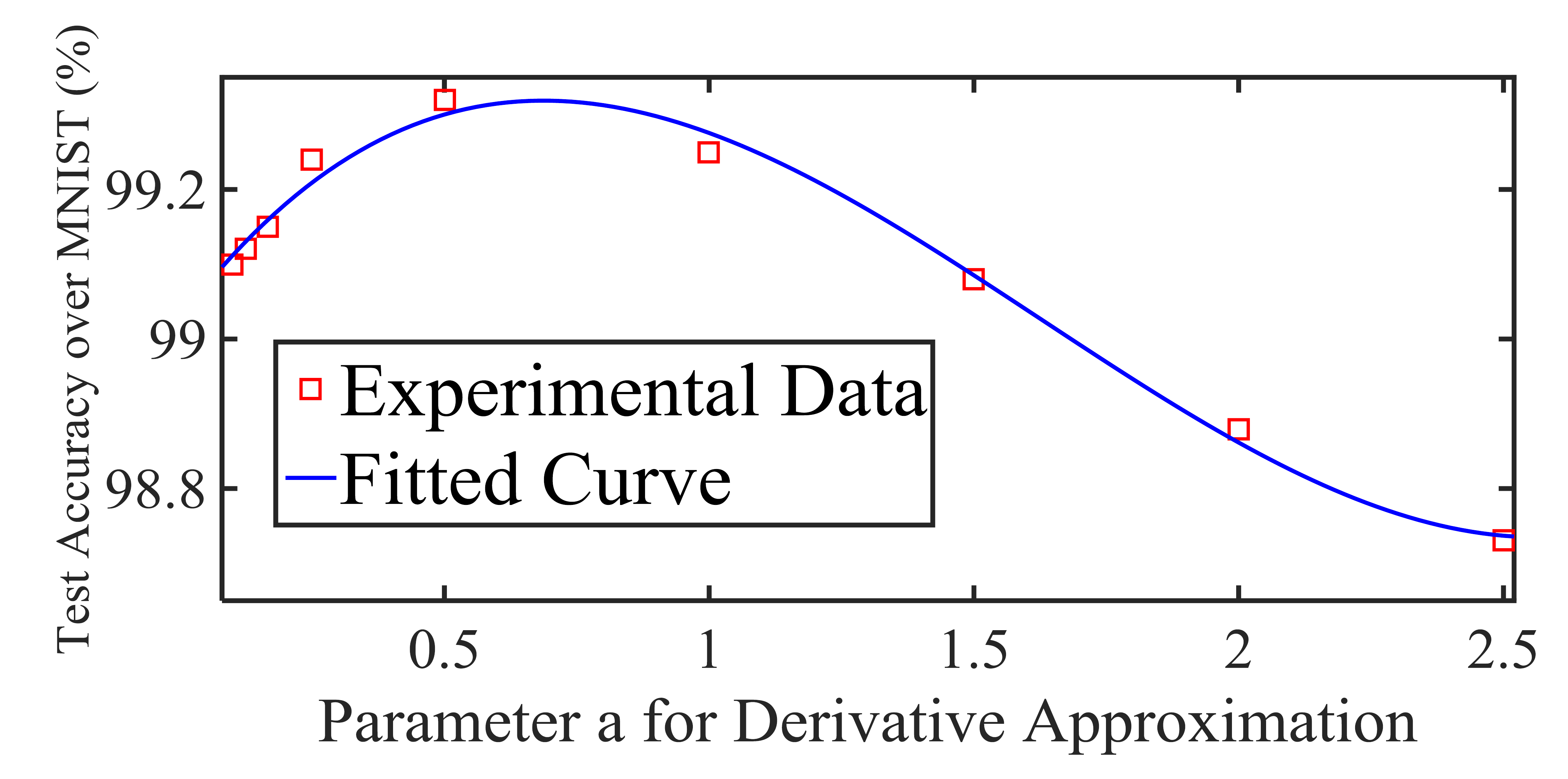}
\caption{\textbf{Influence of the pulse width for derivative approximation.} The pulse width for derivative approximation of non-differentiable discretized activation function affects the network performance. `Not too wide \& not too narrow pulse' achieves the best accuracy.}
\label{derivative_a}
\end{center}
\end{figure}

Secondly, we use the rectangular approximation in Fig. \ref{neuron_discrete1}(c) as an example to explore the impact of pulse width on the recognition performance, as shown in Fig. \ref{derivative_a}. Both too large and too small $a$ value would cause worse performance and in our simulation, $a=0.5$ achieves the highest testing accuracy. In other words, there exists a best configuration for approximating the derivative of non-linear discretized activation function.

\begin{figure}[!htbp]
\begin{center}
\includegraphics[height=4.4cm]{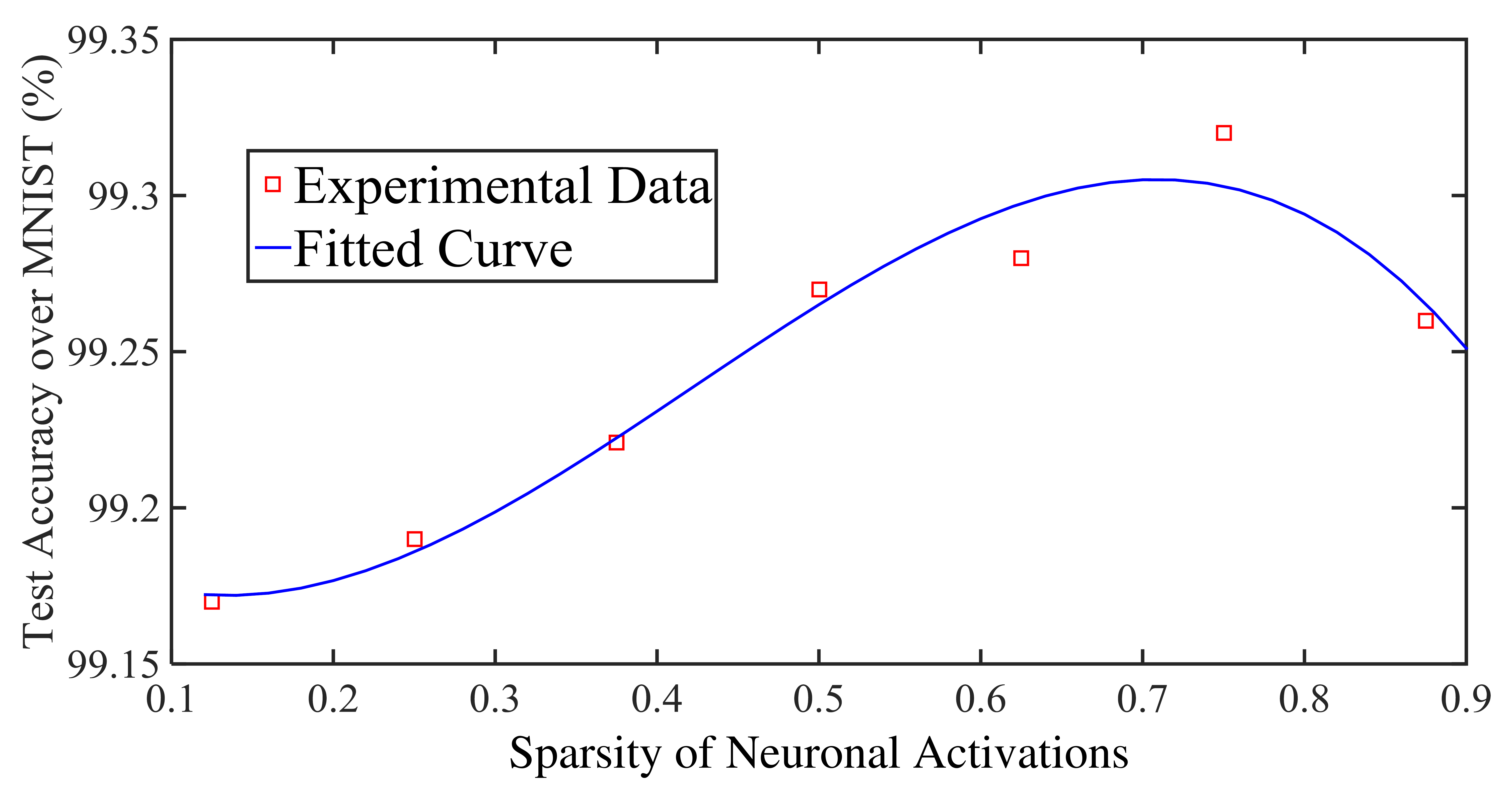}
\caption{\textbf{Influence of the sparsity of neuronal activations.} Here the sparsity represents the fraction of zero activations. By properly increasing the zero neuronal activations, i.e. computation sparsity, the recognition performance can be improved. There exists a best sparse space of neuronal activations for a specific network and dataset.}
\label{sparsity}
\end{center}
\end{figure}

Finally, we investigate the influence of this sparsity on the network performance, and the results are presented in Fig. \ref{sparsity}. Here the sparsity represents the fraction of zero activations. By controlling the width of sparse window (determined by $r$) in Fig. \ref{neuron_discrete1}(a),   the sparsity of neuronal activations can be flexibly modified. It is observed  that the network usually performs better when  the state sparsity properly increases. Actually, the performance significantly degrades when the sparsity further increases, and it approaches  $0$  when the sparsity approaches $1$. This indicates that there exists a best sparse space for a specified network and data set, which is probably due to the fact that the proper increase of zero neuronal activations reduces the network complexity, and the overfitting  can be  avoided to a great extent, like the dropout technology \cite{dropout 2014}. But the valid neuronal information will reduce significantly if the network  becomes  too sparse, which causes the performance degradation. Based on this analysis, it is easily to understand the reason that why the GXNOR-Nets in this paper usually perform better than the BWNs, BNNs and TWNs. On the other side, a sparser network can be more hardware friendly which means that it is possible to achieve higher accuracy and less hardware overhead in the meantime by configuring the computational  sparsity.

\subsection{Event-driven hardware computing architecture}
\begin{figure}[!htbp]
\begin{center}
\includegraphics[height=12cm]{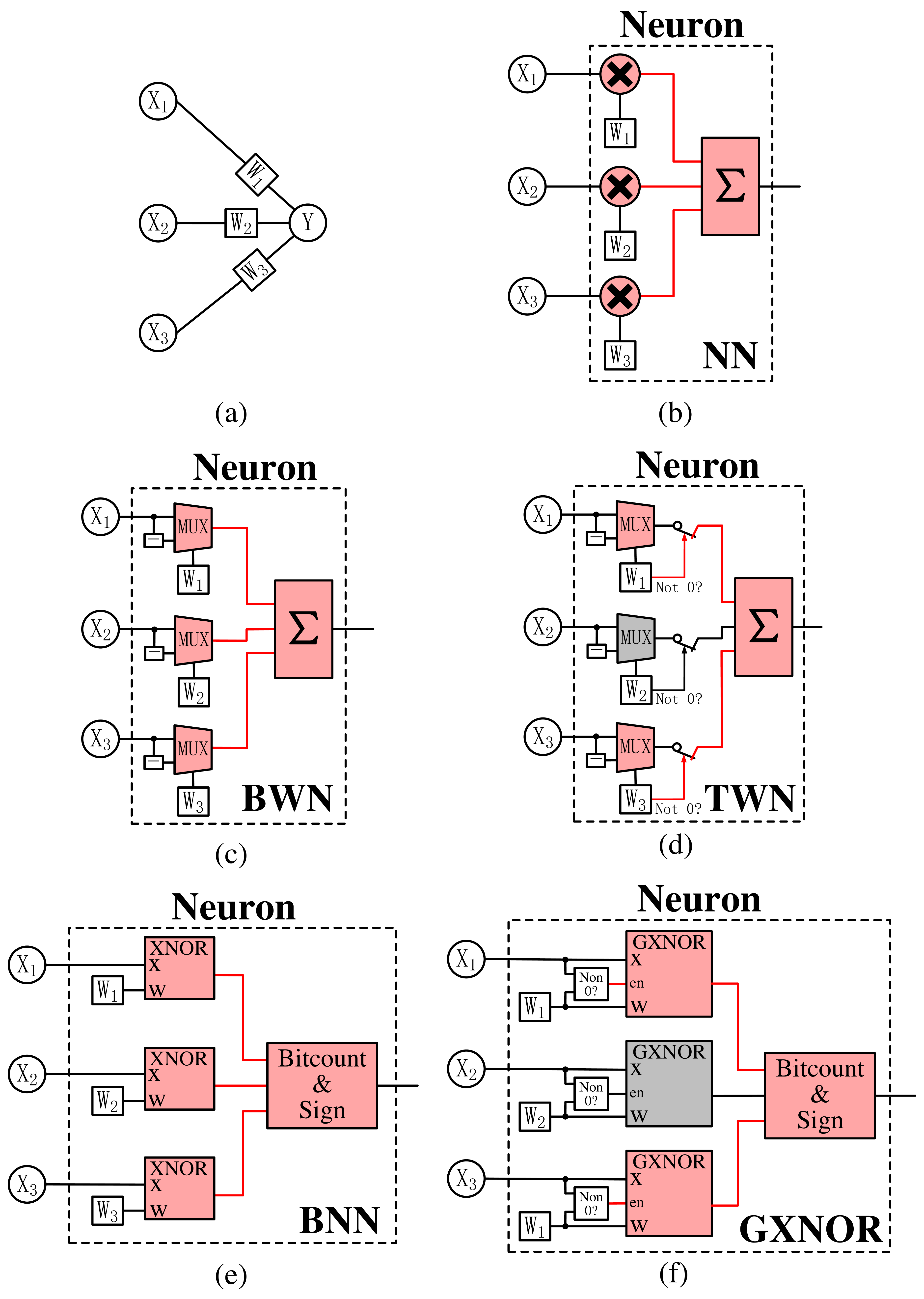}
\caption{\textbf{Comparisons of hardware computing architectures.}  (a) A neural network example of one neuron Y with three inputs $X_1$, $X_2$, $X_3$ and the corresponding synaptic weights $W_1$, $W_2$, $W_3$.  (b) Full-precision Neural Network (NN) with multipliers and an accumulator. (c) Binary Weight Network (BWN) with multiplexers and an accumulator. (d) Ternary Weight Network (TWN) with multiplexers and an accumulator, under event-driven control. (e) Binary Neural Network (BNN) with XNOR and  bitcount operations. (f) GXNOR-Net  with XNOR and bit count operations, under event-driven control.}
\label{computing_architecture}
\end{center}
\end{figure}

\begin{table*}
\centering
\renewcommand\thetable{\arabic{table}}
\caption{\textbf{ Operation overhead comparisons with different computing architectures.}}
\begin{tabular}{ccccccccccc}
\hline
\multirow{2}*{Networks} &\multicolumn{4}{c}{Operations}   &\multirow{2}*{Resting Probability} \\
&\cline{1-4}
 &Multiplication &Accumulation &XNOR &BitCount  \\
 \hline
Full-precision NNs 	&M 	&M  &0 &0  &0.0\%   \\
 \hline
BWNs 	&0 	&M  &0 &0  &0.0\%   \\
 \hline
TWNs 	&0 	&0$\sim$M  &0 &0  &33.3\%   \\
 \hline
BNNs or XNOR Networks 	&0 	&0  &M &1  &0.0\%   \\
 \hline
\textbf{GXNOR-Nets} 	&\textbf{0} 	&\textbf{0}  &\textbf{0$\sim$M} &\textbf{0/1}  &\textbf{55.6\%}   \\
\hline
\label{computing_architecture_table}
\end{tabular}
\end{table*}

\begin{figure*}[!htbp]
\begin{center}
\includegraphics[height=7.5cm]{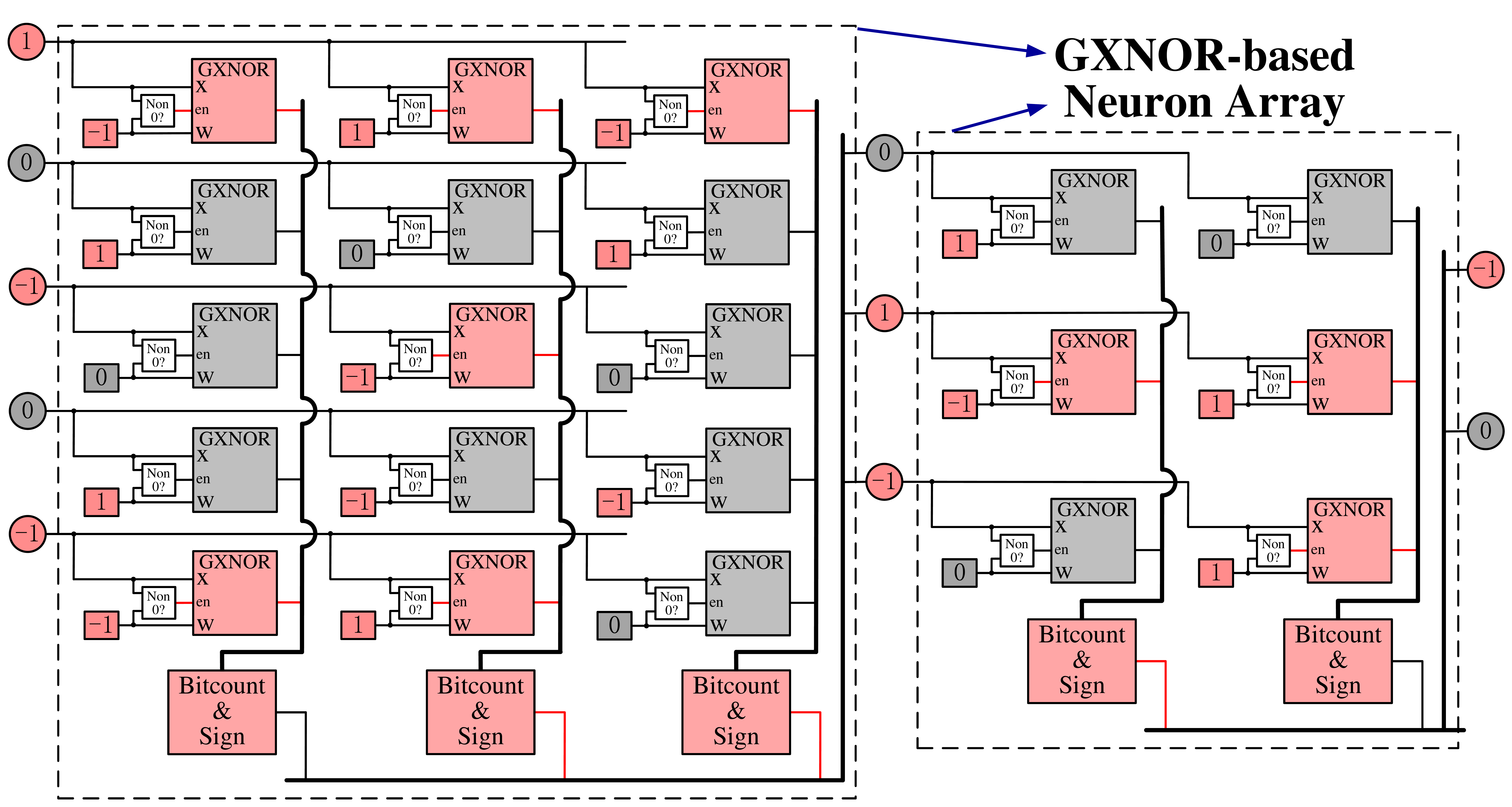}
\caption{\textbf{Implementation of the GXNOR-Net example.} By introducing the event-driven paradigm, most of the operations are efficiently kept in the resting state until the valid gate control signals wake them up. The signal is determined by whether both the weight and activation are non-zero.}
\label{GXNOR_Example}
\end{center}
\end{figure*}

For  the different networks in Table \ref{performance}, the hardware computing architectures can be  quite different. As illustrated in Fig. \ref{computing_architecture}, we present typical hardware implementation examples for a  triple-input-single-output neural network, and the corresponding original network is shown in Fig. \ref{computing_architecture}(a). The conventional hardware implementation for full-precision NN is based on multipliers for the multiplications of activations and weights,  and accumulator for the dendritic integration, as shown in Fig. \ref{computing_architecture}(b). Although  a unit for nonlinear activation function is required,   we ignore this in all cases of Fig. \ref{computing_architecture}, so that we can focus on the influence on the implementation architecture with different discrete spaces. The recent BWN in Fig. \ref{computing_architecture}(c) replaces the multiply-accumulate operations by a simple accumulation operation, with the help of multiplexers. When $W_i=1$, the neuron accumulates $X_i$; otherwise, the neuron accumulates $-X_i$. In contrast, the TWN in Fig. \ref{computing_architecture}(d) implements the accumulation under an event-driven paradigm by adding a zero state into the binary weight space. When $W_i=0$, the neuron is regarded as resting; only when the weight $W_i$ is non-zero, also termed as an event, the neuron accumulation will be activated. In this sense, $W_i$ acts as a control gate. By constraining both the synaptic weights and neuronal activations in the binary space, the BNN in Fig. \ref{computing_architecture}(e) further simplifies the accumulation operations in the BWN to efficient binary logic XNOR and bitcount operations. Similar to the event control of BNN, the TNN proposed in this paper further introduces the event-driven paradigm based on the binary XNOR network. As shown in Fig. \ref{computing_architecture}(f), only when both the weight $W_i$ and input $X_i$ are non-zero, the XNOR and bit count operations are enabled and started. In other words, whether  $W_i$ or  $X_i$ equals to zero or not plays the role of closing or opening of the control gate, hence the name of gated XNOR network (GXNOR-Net) is granted.

Table \ref{computing_architecture_table} shows the required operations of the typical networks in Fig. \ref{computing_architecture}. Here we assume that the input number of the neuron is $M$, i.e. $M$ inputs and one neuron output. We can see that the BWN removes the multiplications in the original full-precision NN, and the BNN replaces the arithmetical operations to efficient XNOR logic operations. While, in full-precision NNs, BWNs (binary weight networks), BNNs/XNOR networks (binary neural networks), most states of the activations and weights are non-zero. So their resting probability is $\approx 0.0\%$. Furthermore, the TWN and GXNOR-Net introduce the event-driven paradigm. If the states in the ternary space $\{-1,0,1\}$ follow uniform distribution, the resting probability of accumulation operations in the TWN reaches 33.3\%, and the resting probability of XNOR and bitcount operations in  GXNOR-Net  further reaches 55.6\%. Specifically, in TWNs (ternary weight networks), the synaptic weight has three states $\{-1, 0, 1\}$ while the neuronal activation is fully precise. So the resting computation only occurs when the synaptic weight is $0$, with average probability of $\frac{1}{3}\approx 33.3\%$. As for the GXNOR-Nets, both the neuronal activation and synaptic weight have three states $\{-1, 0, 1\}$. So the resting computation could occur when either the neuronal activation or the synaptic weight is $0$. The average probability is $1-\frac{2}{3}\times \frac{2}{3}=\frac{5}{9}\approx 55.6\%$. Note that Table 2 is based on an assumption that the states of all the synaptic weights and neuronal activations subject to a uniform distribution.   Therefore the resting probability  varies  from different networks and data sets and  the reported values can only be used as rough guidelines.

Fig.  \ref{GXNOR_Example} demonstrates an example of hardware implementation of the   GXNOR-Net from Fig. \ref{TNN}. The original $21$ XNOR operations can be reduced to only $9$ XNOR operations, and the required bit width for the bitcount operations can also be reduced. In other words, in a GXNOR-Net, most operations keep in the resting state until the valid gate control signals wake them up, determined by whether both the weight and activation are non-zero. This sparse property promises the design of ultra efficient intelligent devices with the help of event-driven paradigm, like the famous event-driven TrueNorth neuromorphic chip from IBM \cite{TrueNorth 2014, TrueNorth 2016}.

\subsection{Multiple states in the discrete space}
\begin{figure}[!htbp]
\begin{center}
\includegraphics[height=5.1cm]{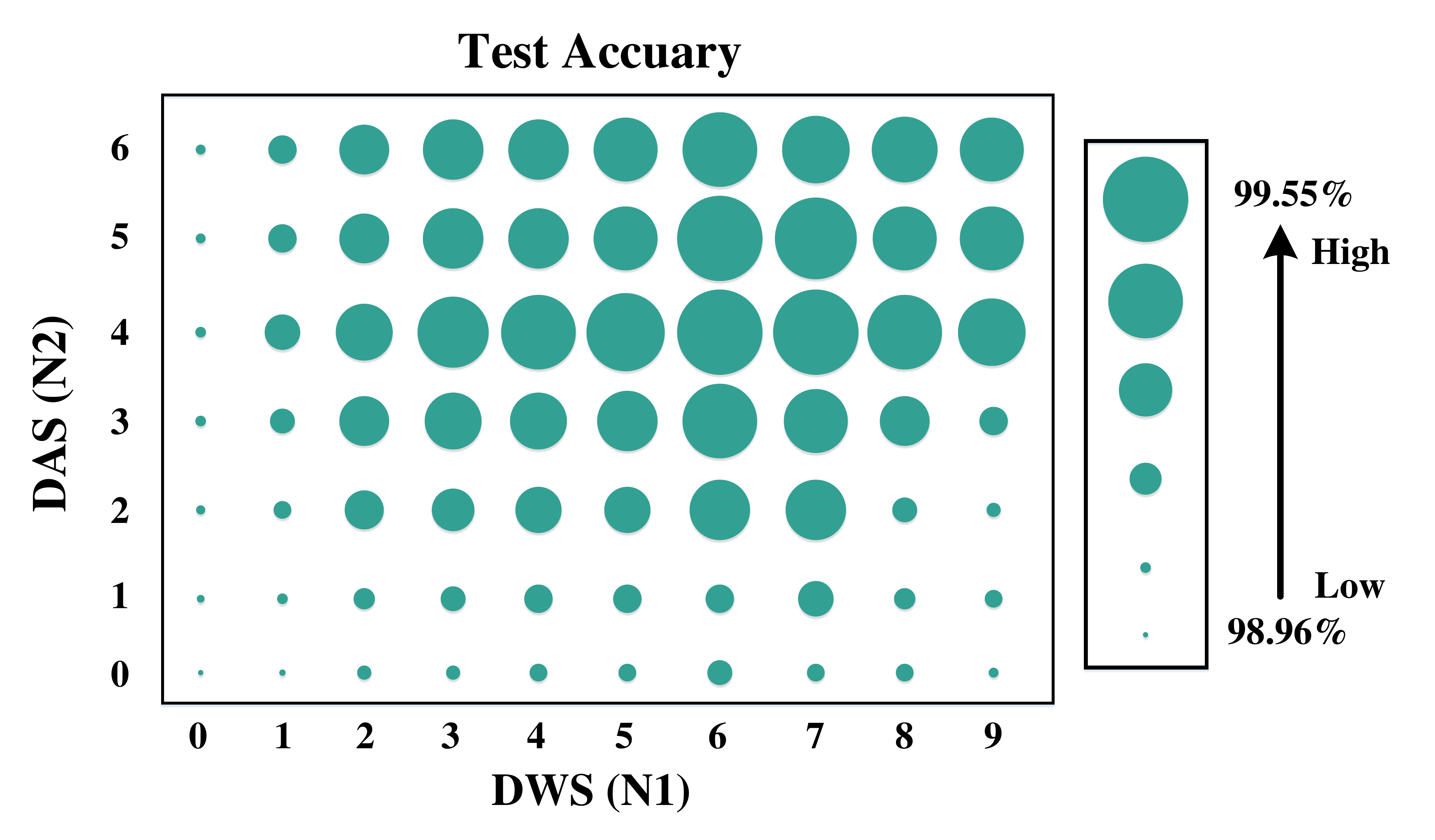}
\caption{\textbf{Influence of the state number in the discrete space.} The state number in discrete spaces of weights and activations can be multi-level values, i.e. DWS and DAS. The state parameters of weight space and activation space are denoted as $N_1$ and $N_2$, respectively, which is similar to the definition of $Z_N$ in (\ref{ZN}). There exists a best discrete space with respect to  either the weight direction or the activation direction, which locates at $N_1=6$ and $N_2=4$.}
\label{multi_level}
\end{center}
\end{figure}
According to  Fig. \ref{neuron_discrete2} and Fig. \ref{weight_discrete3}, we know that the discrete spaces of synaptic weights and neuronal activations can have multi-level states. Similar to the definition of $Z_N$ in (\ref{ZN}), we denote the state parameters of  DWS and DAS as $N_1$ and $N_2$, respectively.  Then, the available state number of weights and activations are $2^{N_1}+1$ and $2^{N_2}+1$, respectively. $N_1=0$ or $N_1=1$ corresponds to binary or ternary weights, and $N_2=0$ or $N_2=1$ corresponds to binary or ternary activations. We test the influence of $N_1$ and $N_2$ over MNIST dataset, and Fig. \ref{multi_level} presents the results where the larger circle denotes higher test accuracy. In the weight direction,  it is observed that when $N_1=6$, the network performs best; while in the activation direction, the best performance occurs when $N_2=4$. This indicates there    exists a best discrete space in either the weight direction or the activation direction, which is similar to the conclusion from the influence analysis of $m$ in Fig. \ref{tanh_m}, $a$ in Fig. \ref{derivative_a}, and sparsity in Fig. \ref{sparsity}. In this sense, the discretization is also an efficient way to avoid network overfitting  that improves  the algorithm performance. The investigation in this section can be used as a guidance theory to help us choose a best discretization implementation for a particular hardware platform after considering its computation and memory resources.

\section{Conclusion and Discussion}
This work  provides  a unified discretization framework for both synaptic weights and neuronal activations in DNNs, where the derivative of multi-step activation function is approximated and the storage of full-precision hidden weights is avoided by using a probabilistic projection operator to directly realize DST. Based on this, the  complete back propagation learning process can be conveniently implemented when both the weights and activations are discrete. In contrast to the existing binary or ternary methods, our model can flexibly modify the state number of weights and activations to make it suitable for various hardware platforms, not limited to the special cases of binary or ternary values. We test our model in the case of ternary weights and activations (GXNOR-Nets)  over MNIST, CIFAR10 and SVHN datesets, and achieve comparable performance with state-of-the-art algorithms. Actually, the non-zero state  of   the weight and activation acts as a control signal to enable   the computation unit, or keep it resting. Therefore GXNOR-Nets can be regarded as one  kind of ``sparse  binary networks" where the networks' sparsity can be controlled through adjusting a pre-given parameter. What's more, this ``gated control'' behaviour promises the design of efficient hardware implementation by using event-driven paradigm, and this has been compared with several typical neural networks and their hardware computing architectures. The  computation sparsity and the number of states  in  the discrete space can be properly increased to further improve the recognition performance of  the GXNOR-Nets.

 We  have also tested the  performance of the  two curves  in Fig. \ref{neuron_discrete1} for derivative approximation.  It is found  that the pulse shape (rectangle or triangle) affect less on the accuracy compared to the pulse width (or steepness) as shown in Fig. \ref{derivative_a}. Therefore, we recommend to use the rectangular one in Fig. \ref{neuron_discrete1}(c) because it is simpler than the triangular curve in Fig. \ref{neuron_discrete1}(d),  which makes  the approximation more hardware-friendly.

 Through above analysis, we know that GXNOR-Net can dramatically simplify the computation in the inference phase and reduce the memory cost in the training/inference phase. However, regarding the training computation, although it can remove the multiplications and additions in the forward pass and remove most multiplications in the backward pass, it causes slower convergence and probabilistic sampling overhead. On powerful GPU platform with huge computation resources, it may be able to cover the overhead from these two issues by leveraging the reduced multiplications. However, on other embedded platforms (e.g. FPGA/ASIC), they require elaborate architecture design.

 Although the GXNOR-Nets promise the event-driven and efficient hardware implementation, the quantitative advantages are not so huge if only based on current digital technology. This is because   the  generation   of the control gate signals   also requires extra overhead. But the power consumption can be reduced to a certain extent because of the less state flips in digital circuits, which can be further optimized by increasing the computation sparsity. Even more promising, some emerging nanodevices have the similar event-driven behaviour, such as gated-control memristive devices \cite{Gated_Memristor 1, Gated_Memristor 2}. By using these devices, the multi-level multiply-accumulate operations can be directly implemented, and the computation is controlled by the event signal injected into the third terminal of a control gate. These characteristics naturally match well with our model with multi-level weights and activations by modifying the  number of  states in the discrete space as well as the event-driven paradigm with flexible computation sparsity. $\\$

 \textbf{Acknowledgment.} The work was partially supported by  National Natural Science Foundation of China (Grant No. 61475080, 61603209),  Beijing Natural Science Foundation (4164086),  and Independent Research Plan of  Tsinghua University  (20151080467).

\end{document}